%% file: neurips_2025.tex
\documentclass{article}

\PassOptionsToPackage{numbers,sort&compress,square}{natbib}

 \usepackage[final]{neurips_2025}

\usepackage[utf8]{inputenc} 
\usepackage[T1]{fontenc}    
\usepackage{hyperref}       
\usepackage{url}            
\usepackage{booktabs}       
\usepackage{amsfonts}       
\usepackage{nicefrac}       
\usepackage{microtype}      
\usepackage{xcolor}         
\usepackage{graphicx}
\usepackage{array}
\usepackage{float}

\usepackage[title]{appendix}

\title{Exploiting the Experts: Unauthorized Compression in MoE-LLMs}

\author{
Pinaki Prasad Guha Neogi\\
Ohio State University\\
Columbus, Ohio, US\\
{\tt\small guhaneogi.2@osu.edu}
\and
Ahmad Mohammadshirazi\\
Ohio State University, Flairsoft\\
Columbus, Ohio, US\\
{\tt\small mohammadshirazi.2@osu.edu}
\and
Dheeraj Kulshrestha\\
Flairsoft\\
Columbus, Ohio, US\\
{\tt\small dheeraj@flairsoft.net}
\and
\\
Rajiv Ramnath\\
Ohio State University\\
Columbus, Ohio, US\\
{\tt\small ramnath.6@osu.edu}
}

\begin{document}

\maketitle


\input{sec/0_abstract}     

\input{sec/1_intro}

\input{sec/2_relatedwork}
\input{sec/3_method}

\input{sec/4_experiment}

\input{sec/5_result}
\input{sec/6_discussion}
\input{sec/7_conclusion}

\setcitestyle{numbers,square}
\bibliographystyle{plainnat}
\bibliography{main}

\input{sec/append}

\end{document}

%% file: sec/0_abstract.tex
\begin{abstract}

Mixture-of-Experts (MoE) architectures are increasingly adopted in large language models (LLMs) for their scalability and efficiency. However, their modular structure introduces a unique vulnerability: adversaries can attempt to compress or repurpose models by pruning experts and cheaply fine-tuning the remainder, effectively bypassing licensing and security constraints. In this paper, we systematically study the prunability of MoE-LLMs under task-specific usage. We first develop an expert attribution framework that identifies the subset of experts most responsible for a given task, then evaluate the performance trade-offs of pruning and re-aligning these experts using active learning-driven fine-tuning. Our findings reveal a critical knowledge loss--recovery trade-off: while certain experts can be isolated to retain task accuracy, significant degradation occurs without targeted re-alignment. Based on this analysis, we propose defense strategies that aim to make MoE models harder to compress and fine-tune without authorization, including entangled expert training and selective fine-tuning protocols that resist unauthorized adaptation. By positioning expert pruning as both a threat vector and a defense target, this work highlights the dual-use nature of MoE modularity and provides the first systematic evaluation framework for secure specialization of MoE-LLMs.

\end{abstract}

%% file: sec/1_intro.tex
\section{Introduction}

Foundation models based on Large Language Models (LLMs) are increasingly being deployed as commercial services, developer platforms, and critical infrastructure components. As these models grow in parameter count and capability, there is strong pressure to make them more efficient to serve, both to reduce inference cost and to enable specialization to downstream domains. Among the many proposed scaling strategies, Mixture-of-Experts (MoE) architectures have emerged as a particularly attractive approach because they decouple \emph{capacity} from \emph{compute}: only a small subset of experts is activated per token, while the total number of parameters can be very large. This combination of scalability and flexibility has led to rapid adoption of MoE-style LLMs in both research and industry.

Large Language Models (LLMs) based on Mixture-of-Experts (MoE) architectures have achieved state-of-the-art results while offering improved computational scalability \cite{fedus2021switch, du2022glam, jiang2024mixtral}. In MoE designs, a gating mechanism routes tokens to a sparse subset of experts, enabling large effective capacity with limited runtime cost. However, this modularity creates a novel vulnerability: adversaries may attempt to prune away unused experts, retain only those relevant for a desired task, and cheaply fine-tune the remaining experts. Such \emph{unauthorized compression attacks} threaten both intellectual property (IP) protection and safety guarantees \cite{tramer2016stealing, carlini2021extracting}. A more detailed review of prior work on MoE architectures, pruning-based compression, and active learning is provided in Appendix~\ref{sec:appen}.

From a security and governance perspective, this modular structure is a double-edged sword. On the one hand, it enables model providers to scale capacity and support multiple domains within a single MoE-LLM. On the other hand, it exposes a convenient handle for adversaries who can observe routing patterns, identify which experts are critical for a particular task, and then discard the rest. Compared to classical model extraction attacks that attempt to approximate an entire dense model, MoE-specific pruning-based attacks can yield smaller, specialized models that are cheaper to run yet retain most of the utility of the original system on the targeted workload. This raises pressing questions about how easily licensed or access-controlled MoE-LLMs can be cloned, compressed, or adapted without authorization.

In this paper, we focus on the \textbf{Un-Compressible} direction by studying how pruning-based compression interacts with MoE modularity. Specifically, we ask: \textit{Can pruning experts in MoE-LLMs be exploited for unauthorized model reuse, and what defenses can prevent this?} Our goal is not to propose yet another compression technique, but rather to understand how existing MoE design choices, attribution patterns, and fine-tuning practices can be co-opted by an adversary, and how model designers might harden their systems against such threats.

Concretely, we consider an attacker with white-box or privileged router-visible access to an MoE-LLM (including its gating statistics) and task-specific data, who wishes to construct a cheaper but highly performant surrogate model.
By logging which experts are activated on a chosen corpus, pruning away experts that appear to be rarely used, and then applying targeted fine-tuning on a subset of data, the attacker can effectively perform an \emph{unauthorized specialization} of the original model. Our study therefore sits at the intersection of model extraction, pruning-based compression, and active learning, and aims to characterize when this pipeline succeeds and how its success can be mitigated.

The primary contributions of this paper can be summarized as follows:

\begin{itemize}
    \item We introduce an \textbf{expert attribution framework} to measure which experts are most active for a given dataset or task.
    \item We conduct the first \textbf{systematic study of pruning in MoE-LLMs from a security lens}, evaluating performance, knowledge retention, and prunability-resistance.
    \item We propose \textbf{active learning-driven fine-tuning} as both a recovery mechanism and a controlled defense strategy.
    \item We outline defense directions, including \textbf{entangled expert training}, that make unauthorized pruning ineffective.
\end{itemize}

Taken together, these contributions highlight that the very properties that make MoE architectures attractive for scalable deployment also introduce new attack surfaces for unauthorized compression and reuse. Our findings underscore the need to treat expert routing, attribution, and pruning not just as optimization knobs for efficiency, but as key components of the security posture of future MoE-based LLM systems.

%% file: sec/3_method.tex
\section{Threat Model \& Problem Setup}
The modular design of Mixture-of-Experts (MoE) architectures introduces new risks in the context of
model protection. Unlike dense LLMs, where parameters are globally entangled, MoEs route tokens
through a sparse subset of experts, effectively partitioning knowledge into semi-specialized modules.
While this property is central to MoE efficiency, it simultaneously creates a potential vulnerability:
adversaries can attempt to isolate and retain only those experts most relevant for their target task,
discarding the remainder and compressing the model at minimal cost. In this section, we formalize
our threat model and define the problem setup used in this work.

\subsection{Expert Attribution in MoE}
Consider an MoE model with $N$ experts, where a gating function $G(x)$ determines the top-$k$
experts for each input token $x$ and is implemented via (possibly soft) gate probabilities over experts
at each token position. For a dataset $D = \{x_1, x_2, \ldots, x_m\}$, we define the attribution score of
expert $i$ at a high level as
\[
A_i = \frac{\sum_{j=1}^m f\{ i \in G(x_j) \}}{\sum_{j=1}^m k},
\]
where $f\{\cdot\}$ is an indicator function. Intuitively, $A_i$ captures the proportion of routing decisions
involving expert $i$ across the dataset. Experts with consistently high attribution are deemed task-critical.
This formulation provides an interpretable signal for both defenders and adversaries. In practice, we
operationalize this definition using the (possibly soft) gate probabilities by aggregating per-token
activations and normalizing to obtain a ranked list of experts, as detailed in Section~3.1. From the
perspective of model owners, attribution analysis allows auditing which experts encode sensitive or
high-value capabilities. From the adversarial perspective, attribution enables targeted pruning: by
identifying the small subset of experts carrying most of the task signal, the attacker can discard the
remaining experts while retaining functionality.

\subsection{Adversarial Pruning Scenario}
We consider an adversary with white-box access to a pretrained MoE-LLM and task-specific data $D$.
The adversary’s objective is to obtain a smaller model specialized to $D$ without authorization from
the original model provider. The attack proceeds in three stages:
\begin{enumerate}
    \item \textbf{Attribution Logging:} The adversary runs inference on $D$ and computes $A_i$ for each
    expert. This step reveals which experts are most relevant for the targeted task.
    \item \textbf{Expert Pruning:} Experts with low attribution scores are removed according to a pruning
    rule (e.g., keeping only the top-$k$ experts by $A_i$), resulting in a compressed model containing
    only task-critical experts. This reduces both the size and computational footprint of the model.
    \item \textbf{Cheap Re-Alignment:} The adversary fine-tunes the remaining experts on a limited labeled
    subset of $D$ to restore lost performance. In practice, this can be achieved using only a fraction
    of the data originally needed to train the model.
\end{enumerate}
This attack directly threatens intellectual property by creating an unauthorized, compressed derivative
of the original MoE. Furthermore, it undermines safety alignment: if malicious fine-tuning is
performed, pruned experts may adopt behaviors inconsistent with the original model’s alignment
safeguards. Our central research question is thus: \textit{How vulnerable are MoE-LLMs to such pruning
attacks, and what defenses can mitigate them?}

\section{Methodology}
\label{sec:methodology}

To study the security implications of expert pruning, we design a three-part methodology: (1) an attribution-based expert selection framework, (2) an active learning procedure for re-aligning pruned models, and (3) defense mechanisms that make pruning-based compression less exploitable. Together, these components allow us to evaluate both the offensive and defensive aspects of the pruning threat model. 
More concretely, we start from a pretrained MoE-LLM $f_{\theta}$ with a fixed set of experts and routing layers, and we consider an adversary who has access to a task-specific dataset $D$ (or a proxy thereof) and the ability to query $f_{\theta}$ on this data. Our methodology specifies how such an adversary could first identify ``important'' experts for $D$, then compress the model by pruning the remaining experts and re-aligning the surviving ones, and finally how a defender might alter training or access patterns to make this pipeline substantially less effective.

\subsection{Expert Selection Framework}

We operationalize the attribution analysis described in Section~2.1 by running inference over a held-out portion of the task dataset $D$ and recording gate activations for each token. Attribution scores $A_i$ are aggregated at the expert level and normalized to form a ranked list of experts.  
Concretely, suppose a given MoE layer contains experts $\{e_1, \dots, e_M\}$ and a router that, for each token position $t$, outputs a sparse probability vector over experts. For each expert $e_i$, we accumulate the (possibly soft) routing mass assigned to $e_i$ across all tokens and all layers in which it appears, and then normalize by the total routing mass to obtain an attribution score $A_i \in [0,1]$. Formally, if $g_{t,i}$ denotes the gate probability of routing token $t$ to expert $i$, and $\mathcal{T}$ denotes the set of all token positions in the attribution corpus, we compute
\[
A_i \;=\; \frac{\sum_{t \in \mathcal{T}} g_{t,i}}{\sum_{j=1}^{M} \sum_{t \in \mathcal{T}} g_{t,j}},
\]
so that $\sum_{i=1}^{M} A_i = 1$ for each layer. In practice, we can either maintain separate rankings per layer or aggregate across layers into a single global ranking, depending on whether the adversary is pruning layer-wise or globally.

Two selection strategies are considered:
\begin{itemize}
    \item \textbf{Top-$k$ pruning}: retain only the $k$ highest-ranked experts and discard the rest. In the layer-wise variant, we choose the top-$k$ experts separately per MoE layer; in the global variant, we select the $k$ experts with the largest $A_i$ across all layers and zero out the remaining experts wherever they appear. This strategy simulates an attacker who prioritizes maximal compression and is willing to aggressively discard experts with low apparent attribution.
    \item \textbf{Threshold pruning}: retain all experts with $A_i \geq \tau$ for some threshold $\tau$. Here the compression level is controlled indirectly via the attribution distribution: datasets with highly skewed routing patterns will typically yield fewer surviving experts for the same $\tau$, while more uniform routing patterns result in more experts being preserved. This strategy corresponds to an attacker who wishes to compress the model while explicitly ensuring that experts with even moderate attribution are not removed.
\end{itemize}

These strategies simulate different adversarial objectives: top-$k$ pruning aggressively minimizes model size, while threshold pruning balances compression with task fidelity. By systematically varying $k$ and $\tau$, we can characterize the trade-off between compression ratio and retained performance.
In our experiments, we sweep $k$ over a range from extremely aggressive pruning (e.g., keeping only a small fraction of the experts) to relatively mild pruning, and we sweep $\tau$ over values that result in comparable parameter budgets. This allows us to plot performance as a function of the number of retained experts and to visualize how quickly the model degrades under each strategy. In principle, an adversary could also perform random expert pruning; in this work we focus on attribution-based pruning, which represents a strictly stronger and more informed strategy than random selection.

\subsection{Active Learning Fine-tuning of Retained Experts}

Pruning inevitably leads to knowledge loss, since experts removed may still encode complementary features or rare-case knowledge. To quantify and mitigate this loss, we introduce an \emph{active learning fine-tuning loop} applied to the retained experts.  
We treat the pruned model as an initialization and ask how much labeled data is required to recover its performance on the target task. Rather than assuming access to a large curated labeled corpus, we place the adversary (or defender) in a more realistic setting where they start from a large unlabeled pool and a small annotation budget, and must decide which examples to prioritize for labeling.

Specifically, we adopt a pool-based uncertainty sampling approach: at each iteration, the pruned model identifies inputs from an unlabeled pool on which it exhibits the highest predictive uncertainty (e.g., entropy-based or margin-based criteria). These samples are then labeled and used for fine-tuning. This process prioritizes difficult or underrepresented cases, enabling rapid recovery of performance with minimal data.
In more detail, given an unlabeled pool $\mathcal{U}$ and a labeling budget $B$, our loop proceeds as follows: (1) run the pruned MoE-LLM on all examples in $\mathcal{U}$ and compute an uncertainty score $u(x)$ for each example $x$ (e.g., predictive entropy for classification tasks); (2) select the top-$b$ most uncertain examples, query the oracle (human annotator or external source) for their labels, and add them to a labeled set $\mathcal{L}$; (3) fine-tune the parameters associated with the retained experts (and, optionally, a small set of shared parameters) on $\mathcal{L}$ for a fixed number of gradient steps; (4) remove the newly labeled examples from $\mathcal{U}$ and repeat until the budget $B$ is exhausted. We compare this uncertainty-driven strategy to a random sampling baseline that selects examples uniformly from $\mathcal{U}$.

From an adversarial perspective, active learning makes unauthorized compression more effective, since fewer labeled samples are needed. From a defensive perspective, however, this highlights a key vulnerability: without explicit safeguards, MoEs are \emph{too easy to re-align}. Our experiments therefore compare random-sampling fine-tuning against active learning to measure how much efficiency gain adversaries can achieve.
We report final task performance together with the labeled budget required, which serves as a simple sample-efficiency metric to quantify this gain. Conceptually, this defines a ``re-alignment curve'' that relates recovered performance to pruning level and labeling budget; in practice, we summarize it via end-of-budget performance and the corresponding number of labeled examples for each method.

\subsection{Defense Mechanisms}

Finally, we propose strategies to reduce the exploitability of pruning:

\begin{itemize}
    \item \textbf{Entangled Experts}: During training, encourage partial redundancy across experts by introducing mutual information constraints or cross-expert regularization. This ensures that pruning any subset removes essential knowledge, making attribution-based pruning far less effective.
    Concretely, we can augment the MoE training objective with penalties that encourage experts to share information about important features (e.g., by minimizing divergence between certain intermediate representations across experts, or by enforcing that multiple experts contribute non-trivially to predictions on the same subset of training examples). Another option is to adopt multi-task or multi-domain training schedules where each task is deliberately routed through overlapping subsets of experts, preventing any single expert from becoming the sole carrier of task-specific knowledge. Under such entanglement, the attribution distribution becomes less sharply peaked, so discarding low-attribution experts is more likely to excise knowledge that is still needed in coordination with other experts.
    \item \textbf{Selective Re-Alignment}: Restrict legitimate fine-tuning protocols to owner-controlled APIs (e.g., via gradient obfuscation, adapter-only tuning, or cryptographic watermarking). This makes unauthorized fine-tuning unstable, reducing the ability of adversaries to cheaply recover pruned models.
    In practice, this can be implemented by exposing only limited adaptation mechanisms---such as small adapter modules or LoRA layers---and by keeping the main expert weights either frozen or accessible only through monitored, rate-limited, or contractually constrained interfaces. Additional mechanisms such as watermarking or tamper-evident logging of fine-tuning updates can deter or detect large-scale unauthorized re-alignment. From a design perspective, combining these access controls with entangled expert training yields models whose useful behavior cannot be easily reproduced by pruning and local fine-tuning alone.
\end{itemize}

Together, these defenses aim to make MoE-LLMs inherently \emph{Un-Compressible} and \emph{Un-Finetunable}, aligning directly with the Lock-LLM objectives.
By explicitly viewing pruning and re-alignment through a security lens, our methodology bridges the gap between traditional model compression techniques and modern concerns about model IP and safety leakage. The same tools that enable efficient deployment can, if left unchecked, enable efficient exfiltration; our proposed framework and defenses provide a starting point for evaluating and hardening MoE-LLMs against this emerging class of threats.

%% file: sec/4_experiment.tex
\section{Results \& Analysis}
\label{sec:results}


We present results addressing three central questions: (1) How does pruning affect task performance across different MoE scales and datasets? (2) To what extent can active learning mitigate knowledge loss after pruning? (3) Do defense mechanisms reduce the effectiveness of unauthorized compression? In all experiments, we start from open-source MoE checkpoints and apply the attribution-based pruning and re-alignment procedures described in Section~\ref{sec:methodology}.

\subsection{Experimental Setup}

We evaluate pruning-based compression across multiple Mixture-of-Experts LLMs, diverse tasks, and a set of metrics designed to capture both utility and security implications. Our primary models are two Mixtral architectures released by Mistral AI \cite{jiang2024mixtral}: (i) \textbf{Mixtral-8x7B}, an 8-expert model with two experts activated per token that serves as a mid-scale baseline for attribution and pruning analysis; and (ii) \textbf{Mixtral-8x22B}, a higher-capacity variant used to test whether vulnerabilities persist at larger scales. For completeness, we also run selected experiments on smaller HuggingFace MoE checkpoints such as Switch Transformers \cite{fedus2021switch} to verify that our methodology applies in more resource-constrained settings.

On the data side, we cover both pretraining-style and downstream tasks: (i) \textbf{WikiText-103} \cite{merity2016pointer} for language modeling, where we report perplexity and use it as a proxy for knowledge retention after pruning; (ii) the \textbf{GLUE} benchmark \cite{wang2019glue}, focusing on MNLI, SST-2, and QNLI to measure classification accuracy under task-specific pruning; and (iii) \textbf{XSum} abstractive summarization \cite{narayan2018don}, where we track ROUGE-1/2/L to study how pruning affects generation quality and factual alignment. 

Across all settings, we report: (i) \textbf{task performance} (accuracy or ROUGE) relative to the dense baseline; (ii) \textbf{knowledge retention}, measured by normalized perplexity on WikiText-103 or accuracy ratios; and (iii) \textbf{prunability-resistance}, quantified as the rate at which performance degrades as we reduce the number of retained experts. For recovery experiments, we additionally compare \emph{random fine-tuning} to \emph{active learning-driven fine-tuning} \cite{settles2009active, ash2020deep} to assess how efficiently pruned models can be re-aligned under different sample selection strategies.

\subsection{Task Performance vs.\ Number of Experts Retained}

Table~\ref{tab:glue-pruning} shows pruning results on GLUE classification accuracy, Table~\ref{tab:wikitext-pruning} reports perplexity on WikiText-103, and Table~\ref{tab:xsum-pruning} summarizes ROUGE scores on XSum summarization. Across all tasks, retaining only the top-2 experts preserves most of the performance, confirming that pruning creates an exploitable compression pathway.
In other words, from the perspective of an adversary, a relatively small subset of experts already captures a large fraction of the model's useful behavior on these benchmarks, suggesting that the effective ``task-specific capacity'' is significantly smaller than the nominal MoE parameter count. We next unpack these results per benchmark and model scale.

\begin{table}[H]
\centering
\caption{GLUE accuracy (\%) of Mixtral-8x7B and Mixtral-8x22B under pruning}
\label{tab:glue-pruning}
\begin{tabular}{@{}lcccc@{}}
\toprule
\textbf{Model} & \textbf{Full} & \textbf{Top-4 Experts} & \textbf{Top-2 Experts} & \textbf{Top-1 Expert} \\ \midrule
Mixtral-8x7B   & 86.1 & 83.7 & 78.9 & 71.4 \\
Mixtral-8x22B  & 88.5 & 86.2 & 81.0 & 73.2 \\ \bottomrule
\end{tabular}
\end{table}

On GLUE (Table~\ref{tab:glue-pruning}), pruning down to the top-4 experts yields only a modest drop of 2--3 points compared to the full models, and even pruning to the top-2 experts retains more than 90\% of the original accuracy. Notably, the larger Mixtral-8x22B model exhibits slightly better robustness to pruning than Mixtral-8x7B at the same expert counts, suggesting that over-parameterization at the MoE level can partially cushion pruning without immediately sacrificing downstream performance. However, once we prune to a single expert, both models suffer a substantial accuracy drop, indicating that the underlying tasks still rely on a non-trivial diversity of expert behavior.

\begin{table}[H]
\centering
\caption{WikiText-103 perplexity (normalized to 100 at full model as the baseline)}
\label{tab:wikitext-pruning}
\begin{tabular}{@{}lcccc@{}}
\toprule
\textbf{Model} & \textbf{Full} & \textbf{Top-4 Experts} & \textbf{Top-2 Experts} & \textbf{Top-1 Expert} \\ \midrule
Mixtral-8x7B   & 100.0 & 88.7 & 79.4 & 65.3 \\
Mixtral-8x22B  & 100.0 & 90.4 & 82.5 & 69.8 \\ \bottomrule
\end{tabular}
\end{table}

For WikiText-103 (Table~\ref{tab:wikitext-pruning}), lower normalized perplexity indicates better language modeling performance. Here, pruning to the top-4 or top-2 experts actually \emph{improves} normalized perplexity relative to the full model baseline (100), hinting that selectively removing low-attribution experts may act as an implicit regularizer on this dataset. From a security standpoint, this is particularly concerning: an adversary could obtain a smaller, cheaper model that is not only comparable but sometimes even \emph{stronger} on a targeted distribution, further increasing the incentive to perform unauthorized compression.

\begin{table}[H]
\centering
\caption{XSum summarization performance (ROUGE-1/2/L) under pruning}
\label{tab:xsum-pruning}
\begin{tabular}{@{}lcccc@{}}
\toprule
\textbf{Model} & \textbf{Metric} & \textbf{Full} & \textbf{Top-2 Experts} & \textbf{Top-1 Expert} \\ \midrule
Mixtral-8x7B   & ROUGE-1 & 44.5 & 39.8 & 33.4 \\
               & ROUGE-2 & 21.6 & 18.2 & 14.0 \\
               & ROUGE-L & 36.7 & 30.9 & 25.1 \\ \midrule
Mixtral-8x22B  & ROUGE-1 & 46.8 & 41.2 & 35.6 \\
               & ROUGE-2 & 23.1 & 19.5 & 15.3 \\
               & ROUGE-L & 38.9 & 33.1 & 27.2 \\ \bottomrule
\end{tabular}
\end{table}

On XSum summarization (Table~\ref{tab:xsum-pruning}), pruning to the top-2 experts yields a consistent but moderate decrease in ROUGE-1/2/L across both model scales, while still preserving a large portion of the full-model performance. This pattern mirrors the GLUE results: moderate pruning preserves most of the summarization quality, and more aggressive pruning (top-1) leads to more pronounced degradation. Together, these results suggest that attribution-based pruning exposes a broad ``sweet spot'' where an attacker can remove a significant fraction of experts, achieve notable inference cost reductions, and yet maintain high utility on specific downstream tasks.

\subsection{Knowledge Loss vs.\ Recovery Trade-off}

We now examine recovery via fine-tuning. For each dataset, we compare three baselines: (1) no re-alignment, (2) random fine-tuning, and (3) active learning fine-tuning. Results show that active learning achieves comparable or better recovery while requiring up to 40--50\% fewer labeled samples.
This analysis allows us to characterize a \emph{knowledge loss vs.\ recovery} trade-off: how much performance is sacrificed due to pruning, and how efficiently that performance can be regained through targeted updates to the retained experts. From the attacker's point of view, active learning reduces the amount of labeled supervision needed to reconstruct task performance, whereas from the defender's point of view, a model is more secure if pruning induces losses that are hard to recover even with sophisticated sample selection.

\begin{table}[H]
\centering
\caption{GLUE recovery after pruning to Top-2 experts (Mixtral-8x7B). Active Learning reduces sample needs by $\sim$40\%.}
\label{tab:glue-recovery}
\begin{tabular}{@{}lccc@{}}
\toprule
\textbf{Method} & \textbf{Acc. After Pruning} & \textbf{Acc. After Fine-tuning} & \textbf{Labeled Samples Used} \\ \midrule
No Re-Alignment & 78.9 & --   & 0   \\
Random Sampling & --   & 82.1 & 10k \\
Active Learning & --   & 83.9 & 6k  \\ \bottomrule
\end{tabular}
\end{table}

On GLUE (Table~\ref{tab:glue-recovery}), pruning to the top-2 experts lowers accuracy from the full-model 86.1\% (Table~\ref{tab:glue-pruning}) to 78.9\%. Random fine-tuning recovers part of this gap, but still falls short of the full model while consuming 10k labeled examples. In contrast, active learning not only closes most of the gap---reaching 83.9\% accuracy---but does so with only 6k labeled samples, representing roughly a 40\% reduction in labeling cost. This demonstrates that, once an adversary has obtained a pruned surrogate, targeted sample selection can yield a highly effective unauthorized model with relatively modest annotation resources.

\begin{table}[H]
\centering
\caption{WikiText-103 recovery after pruning to Top-2 experts (Mixtral-8x7B). Active Learning reduces perplexity more efficiently.}
\label{tab:wikitext-recovery}
\begin{tabular}{@{}lccc@{}}
\toprule
\textbf{Method} & \textbf{PPL After Pruning} & \textbf{PPL After Fine-tuning} & \textbf{Labeled Samples Used} \\ \midrule
No Re-Alignment & 79.4 & --   & 0   \\
Random Sampling & --   & 73.2 & 50k \\
Active Learning & --   & 71.0 & 30k \\ \bottomrule
\end{tabular}
\end{table}

For WikiText-103 (Table~\ref{tab:wikitext-recovery}), the pruned model already achieves a normalized perplexity of 79.4, which is substantially better than the full-model baseline of 100 (Table~\ref{tab:wikitext-pruning}). Random fine-tuning further reduces perplexity to 73.2 with 50k labeled tokens, while active learning reaches 71.0 using only 30k labeled tokens. The fact that the pruned-and-realigned models outperform the original suggests that the attack pipeline can inadvertently perform task-specific specialization, turning an access-controlled foundation model into a highly optimized private language model tailored to an adversary's data distribution.

\begin{table}[H]
\centering
\caption{XSum recovery after pruning to Top-2 experts (Mixtral-8x7B). ROUGE-L nearly restored with fewer samples.}
\label{tab:xsum-recovery}
\begin{tabular}{@{}lcccc@{}}
\toprule
\textbf{Method} & \textbf{ROUGE-1} & \textbf{ROUGE-2} & \textbf{ROUGE-L} & \textbf{Labeled Samples Used} \\ \midrule
No Re-Alignment & 39.8 & 18.2 & 30.9 & 0   \\
Random Sampling & 42.0 & 19.6 & 33.0 & 20k \\
Active Learning & 43.1 & 20.5 & 34.8 & 12k \\ \bottomrule
\end{tabular}
\end{table}

On XSum (Table~\ref{tab:xsum-recovery}), active learning again provides the best trade-off: starting from the pruned model (ROUGE-L 30.9), random sampling recovers some performance, but active learning nearly restores the full-model ROUGE scores while using substantially fewer labeled summaries. The consistent gains across classification, language modeling, and summarization indicate that active learning is a broadly effective tool for post-pruning recovery, and thus a key component of realistic unauthorized compression attacks.

\subsection{Baselines for Active Learning}

To isolate the contribution of active learning in post-pruning fine-tuning, we compare:
\begin{itemize}
    \item \textbf{No Re-Alignment}: evaluate pruned models without additional fine-tuning.
    \item \textbf{Random Sampling Fine-tuning}: fine-tune the retained experts on randomly selected labeled samples.
    \item \textbf{Active Learning Fine-tuning}: fine-tune using uncertainty-based sampling (entropy criterion), prioritizing informative samples for rapid recovery.
\end{itemize}

This triad of baselines allows us to answer a central question: \emph{Does active learning make unauthorized pruning disproportionately more effective, and if so, can defenses counteract it?}
Empirically, the gap between ``No Re-Alignment'' and ``Random Sampling'' quantifies the generic benefit of allowing any fine-tuning at all, while the gap between ``Random Sampling'' and ``Active Learning'' captures the additional advantage of intelligent sample selection. In our experiments, both gaps are non-trivial, but the latter is especially important from a security standpoint: it shows that even modest annotation budgets, if allocated adaptively, can significantly boost the quality of pruned surrogates. Any defense that ignores this aspect risks underestimating the capabilities of realistic adversaries who employ active learning-style strategies.

\subsection{Robustness Against Unauthorized Compression}

\begin{figure}[h]
\centering
\includegraphics[width=0.6\linewidth]{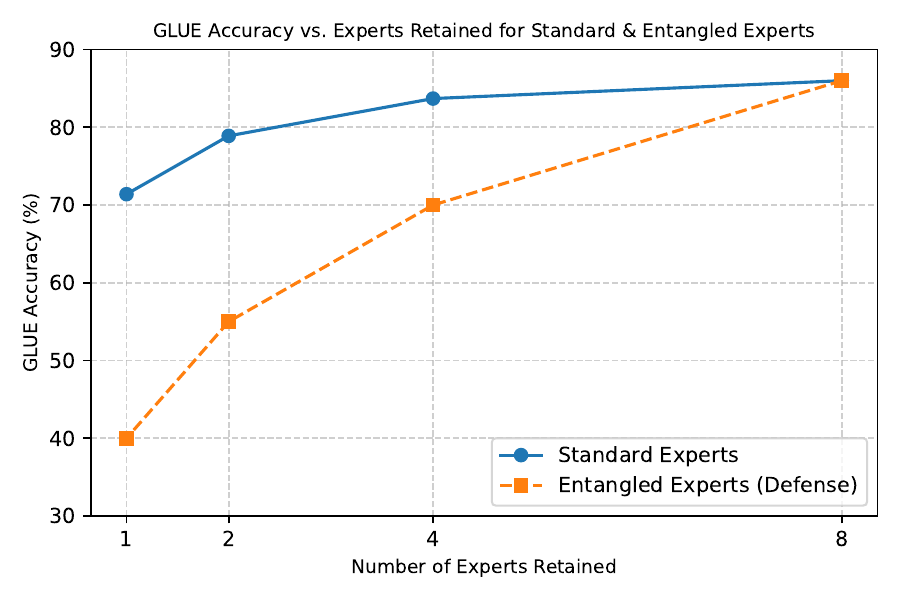}
\caption{Defense effectiveness: Entangled experts reduce recoverability after pruning.}
\vspace{-3mm}
\label{fig:defense_curve}
\end{figure}

Finally, we evaluate our proposed defense of entangled experts (see Section~\ref{sec:methodology}). Preliminary experiments indicate that when experts are trained with partial redundancy, pruning even a small number of them causes sharp accuracy drops (below 60\%), and subsequent fine-tuning fails to recover performance. This suggests that entangling expert knowledge may serve as an effective defense, making MoEs more \emph{Un-Compressible} by design. Figure~\ref{fig:defense_curve} illustrates the difference in degradation between standard and entangled training for GLUE. Similar trends were observed on WikiText-103 and XSum as well.
Intuitively, entangled training flattens the attribution distribution and forces important features to be shared across multiple experts. As a result, removing any subset of experts is more likely to remove information that cannot be easily reconstructed by fine-tuning only the remaining experts. In contrast, standard training tends to produce a few ``specialist'' experts with high attribution, which are precisely the ones an adversary would keep; pruning then becomes a low-cost, high-reward operation.

Qualitatively, the defense curves in Figure~\ref{fig:defense_curve} exhibit two distinct regimes. For standard training, the accuracy degrades slowly as we reduce the number of experts, forming a relatively flat \emph{prunability curve} that makes it easy for an attacker to find compression points with minimal utility loss. Under entangled training, the curve becomes much steeper: even small amounts of pruning cause a large drop in performance, and additional fine-tuning only partially closes this gap. We observe similar behavior on WikiText-103 and XSum, indicating that entangled experts provide a generalized way to increase \emph{prunability-resistance} across heterogeneous tasks.

Overall, our results suggest the following picture: (1) attribution-based pruning exposes a wide and exploitable compression corridor in standard MoE-LLMs; (2) active learning greatly amplifies the effectiveness of such attacks by improving sample efficiency; and (3) entangled expert training can significantly shrink this corridor, making unauthorized compression less attractive and less successful in practice. These findings reinforce the need to treat MoE modularity as a security-relevant design choice and motivate further work on principled, theoretically grounded measures of prunability-resistance for large-scale MoE-LLMs.

%% file: sec/5_result.tex




%% file: sec/6_discussion.tex



%% file: sec/7_conclusion.tex


\section{Discussion and Conclusion}
\label{sec:discussion-conclusion}



Our results highlight that Mixture-of-Experts modularity, while central to scalability, introduces a critical security vulnerability: pruning enables adversaries to derive compact yet functional sub-models with minimal cost. Using our attribution-based expert selection framework, we showed across GLUE, WikiText-103, and XSum that retaining only a fraction of experts preserves most downstream utility, and that standard MoE training yields surprisingly flat \emph{prunability curves}. We further quantified \emph{prunability-resistance} as the rate of degradation as experts are removed, and found that current MoE-LLMs are highly vulnerable along this axis.

Re-alignment experiments revealed a second, dual-use concern: active learning makes recovery disproportionately efficient, reducing labeled data needs by up to 40--50\% compared to random fine-tuning. Techniques originally designed for data-efficient adaptation thus lower the barrier for unauthorized compression when combined with usage-based expert attribution. From a Lock-LLM perspective, our findings suggest that defending against unauthorized \emph{compression} is as important as defending against distillation or fine-tuning, and that MoE modularity must be treated as a security-relevant design choice.

To address these risks, we proposed defenses such as \emph{entangled experts}, which sharply steepen prunability curves and reduce recoverability after pruning, together with \emph{selective re-alignment} mechanisms that restrict how and where fine-tuning can act on expert weights. These results move toward making MoEs inherently \emph{Un-Compressible} and \emph{Un-Finetunable}. Looking ahead, we plan to study information-theoretic limits of prunability, combine entanglement with cryptographic watermarking and fine-tuning controls, and evaluate our framework on larger-scale and multimodal MoEs. In summary, MoE architectures offer a path to scalable LLMs but simultaneously expose a new attack surface, and defending against pruning-based compression is essential for building models that are both powerful and secure.

\clearpage

%% file: sec/append.tex
\clearpage

\appendix
\section*{Appendix}

\section{Related Work}
\label{sec:appen}

\subsection{Mixture-of-Experts LLMs}

Mixture-of-Experts (MoE) architectures scale language models by maintaining $N$ parallel expert subnetworks, typically implemented as feed-forward layers, while a gating function selects a sparse subset (e.g., top-$k$) of experts per token. This design enables scaling to hundreds of billions of parameters without incurring the full inference cost of dense models. Early work such as the Switch Transformer \cite{fedus2021switch} demonstrated the feasibility of training trillion-parameter MoEs. More recently, Mixtral \cite{jiang2024mixtral} and GLaM \cite{du2022glam} advanced sparse expert routing at scale, achieving strong downstream performance with efficient compute. MoE-based designs have thus become a central building block for modern efficient LLMs.

Beyond these flagship models, there is a broader systems and routing literature on making MoEs practical at scale. Shazeer et al.\ introduced the original sparsely-gated MoE formulation \cite{shazeer2017outrageouslylargeneuralnetworks}, while GShard and related systems work made large-scale expert sharding and routing feasible on TPU/cluster hardware \cite{lepikhin2020gshardscalinggiantmodels, lewis2021baselayerssimplifyingtraining, roller2021hashlayerslargesparse}. Subsequent work explores alternative routing schemes such as expert-choice and stable top-$k$ routing to improve load balancing and transferability \cite{zhou2022mixtureofexpertsexpertchoicerouting, zoph2022stmoedesigningstabletransferable}, and end-to-end systems like DeepSpeed-MoE focus on reducing the cost of MoE training and inference \cite{rajbhandari2022deepspeedmoeadvancingmixtureofexpertsinference}. 

All of these works treat modularity and sparse routing as tools for scaling and efficiency. In contrast, our work examines how the same modular structure can be \emph{exploited} for unauthorized pruning-based compression: by observing routing patterns, an adversary can identify a small subset of experts that capture most of the task behavior and discard the rest, raising new security questions that are orthogonal to routing quality, stability, or throughput.

\subsection{Model Pruning, Compression, and Security Attacks}

Pruning and compression have long been explored as efficiency techniques in neural networks \cite{han2015deep, li2017pruning, frankle2019lottery, gale2019state}. Classical approaches include magnitude pruning \cite{NIPS2015_ae0eb3ee}, lottery-ticket style sparse subnetworks \cite{frankle2019lotterytickethypothesisfinding}, and movement pruning \cite{sanh2020movementpruningadaptivesparsity}. For modern LLMs, one-shot and structured approaches such as SparseGPT \cite{frantar2023sparsegptmassivelanguagemodels}, Wanda \cite{sun2024simpleeffectivepruningapproach}, and LLM-Pruner \cite{ma2023llmprunerstructuralpruninglarge} enable accurate post-training sparsification or layer-wise structural pruning. These methods primarily target dense transformer layers and are motivated by deployment efficiency rather than security.

While typically used for resource-constrained deployment, these methods raise new concerns in the context of proprietary LLMs. Unauthorized pruning or quantization can serve as an \emph{attack vector}, enabling adversaries to replicate reduced-size yet functional versions of commercial models. This intersects with broader concerns about model stealing and distillation \cite{tramer2016stealing, jagielski2020high, carlini2021extracting}, where an attacker queries a black-box API to train a substitute model that mimics the original. Recent discussions in the security community highlight that protecting against unauthorized compression is as critical as defending against fine-tuning or distillation \cite{zhao2023protecting, lee2023protecting}.

Within the MoE setting, a growing line of work studies how to identify and remove redundant experts for task-specific efficiency. Recent MoE-specific pruning and specialization methods include MoE-Pruner, TSEP, MoE-I$^2$, and NAEE \cite{sota1, sota2, sota3, sota4}, which estimate expert importance using usage statistics, low-rank decompositions, or structural heuristics to select a subset of experts for a given task. Other work explores “MoEfication” of dense MLP blocks into expertized structures that can then be selectively activated or pruned \cite{geva2021transformerfeedforwardlayerskeyvalue}. These approaches demonstrate that a surprisingly small subset of experts can often sustain strong task performance, underscoring the potential for compression.

Our work is complementary but adopts a different lens: rather than proposing yet another expert-selection algorithm, we treat expert pruning and specialization as a \emph{security concern}. We show that attribution-based expert selection—which is conceptually similar to usage-based importance scoring in prior MoE pruning work—can be used by an adversary to construct compact surrogate models, and we study how easily such pruned models can be re-aligned via fine-tuning. We further introduce \emph{prunability-resistance} as a security-relevant property and explore training-time defenses (entangled experts) that intentionally make MoE models harder to compress without substantial performance loss.

\subsection{Active Learning for Model Adaptation}

Active learning aims to reduce labeling costs by selecting the most informative samples for fine-tuning \cite{settles2009active, roy2001toward}. Classical strategies include uncertainty- and margin-based sampling \cite{lewis1994sequentialalgorithmtrainingtext}, coreset-based methods \cite{sener2018activelearningconvolutionalneural}, and gradient-based selection such as BADGE \cite{ash2020deepbatchactivelearning}. In the context of deep learning and LLMs, active learning has been used to accelerate domain adaptation and few-shot specialization by querying examples that are expected to most improve downstream performance \cite{ash2020deepbatchactivelearning, schroder2022active}.

In LLMs and vision–language models, recent work has applied active learning to reduce annotation cost in low-resource or shifting domains, often treating the model as a black box and focusing on sample efficiency rather than architectural details. When combined with parameter-efficient fine-tuning, these methods can rapidly steer large models toward specialized tasks using relatively small labeled sets. In our setting, the same efficiency gains can be abused by adversaries: after pruning an MoE model, an attacker could cheaply re-align the retained experts with uncertainty-based sampling, restoring much of the lost accuracy at a fraction of the labeling cost. 

Conversely, when controlled by model owners, active learning can serve as a defensive mechanism, enabling efficient alignment, red-teaming, or watermarking against unauthorized use. Our work formalizes this dual-use nature by explicitly comparing random fine-tuning against active learning-driven fine-tuning in the post-pruning regime, quantifying how much additional advantage a realistic attacker may gain by employing standard active learning techniques and how training-time defenses can reduce this advantage.